\documentclass[letterpaper, 10 pt, conference]{ieeeconf}  % Comment this line out if you need a4paper

\IEEEoverridecommandlockouts                              % This command is only needed if
\usepackage{graphics} % for pdf, bitmapped graphics files
\usepackage{epsfig} % for postscript graphics files
\usepackage{mathptmx} % assumes new font selection scheme installed
\usepackage{times} % assumes new font selection scheme installed
\usepackage{amsmath} % assumes amsmath package installed
\usepackage{amssymb}  % assumes amsmath package installed
\usepackage{verbatim}
\usepackage{algorithm}
\usepackage{algorithmic}
\usepackage{color}
\usepackage{wrapfig}
\usepackage{url}
\usepackage{multirow}
\DeclareMathAlphabet{\mathcal}{OMS}{cmsy}{m}{n}

\long\def\ignorethis#1{}

\renewcommand{\eqref}[1]{Equation~(\ref{#1})}

\newcommand{\model}{\mathcal{M}}
\newcommand{\action}{\mathbf{a}}
\newcommand{\state}{\mathbf{x}}
\newcommand{\image}{I_t}
\newcommand{\flow}{\hat{F}}
\newcommand{\pixel}{d} % the designated pixel
\newcommand{\pixelstate}{s} % the state of the desginated pixel
\newcommand{\goal}{g}

\newcommand{\specialcell}[2][c]{%
\begin{tabular}[#1]{@{}l@{}}#2\end{tabular}
}

\title{\LARGE \bf
Deep Visual Foresight for Planning Robot Motion
}
\author{Chelsea Finn$^{1,2}$ and Sergey Levine$^{1,2}$% <-this % stops a space
\thanks{$^{1}$Google Brain, Mountain View, CA}%
\thanks{$^{2}$Berkeley Artificial Intellgence Research (BAIR), Department of Electrical Engineering and Computer Science, University of California, Berkeley, Berkeley, CA 94720}%
}

\begin{document}

\maketitle
\thispagestyle{empty}
\pagestyle{empty}

%%%%%%%%%%%%%%%%%%%%%%%%%%%%%%%%%%%%%%%%%%%%%%%%%%%%%%%%%%%%%%%%%%%%%%%%%%%%%%%%
\begin{abstract}

A key challenge in scaling up robot learning to many skills and environments is removing the need for human supervision, so that robots can collect their own data and improve their own performance
without being limited by the cost of requesting human feedback.
Model-based reinforcement learning holds the promise of enabling an agent to learn to predict the effects of its actions, which could provide flexible predictive models for a wide range of tasks
and environments, without detailed human supervision. We develop a method for combining deep action-conditioned video prediction models with model-predictive control
that uses entirely unlabeled training data. Our approach does not require a calibrated camera, an instrumented training set-up, nor precise sensing and actuation.
Our results show that our method enables a real robot to perform nonprehensile manipulation -- pushing objects -- and can handle novel objects not seen during training.

\end{abstract}

%%%%%%%%%%%%%%%%%%%%%%%%%%%%%%%%%%%%%%%%%%%%%%%%%%%%%%%%%%%%%%%%%%%%%%%%%%%%%%%%
\section{Introduction}
\label{introduction}

Most standard robotic manipulation systems consist of a series of modular components for perception and prediction that can be used to plan actions for handling objects.
Imagine that a robot needs to push a cup of coffee across the table to give it to a human.
This task might involve segmenting an observed point cloud into objects, fitting a 3D model to each object segment,
executing a physics simulator using the estimated physical properties of the cup and, finally, choosing the actions which move the cup to the desired location.
However, when robots encounter previously unseen objects in complex, unstructured environments, this pipelined model-based approach can break down when any one stage has
a sufficiently large modeling error.
In particular, errors early in the process cause compounding errors
later in the process, which in turn can produce actions that are ineffective in the real world: even a small error in the estimated liquid content in the cup or the friction coefficient might cause
the robot to push it too high above its center of mass, causing the contents to spill.
This is the essence of the unstructured open world problem: when the
robot has to deal with the variability of the real world, methods based on rigid hand-engineered processes tend to suffer at the hands of special cases, exceptions,
and unmodeled effects.  % e.g. like novel objects.

% TODO - mention that current learning-based approaches to robotics require information for learning, like a reward function, . In comparison, our learning phase involves extremely minimal human involvement.

Learning-based methods have shown remarkable effectiveness in handling complex, unstructured environments in passive
computer vision tasks, such as image classification~\cite{iv4-siv-16} and object detection~\cite{rfcn-dlhs-16}, by co-adapting low and high-level feature representations.
Motivated by large-scale, unsupervised robotic learning, we consider the question of whether it is possible to replace the hand-engineered robotic manipulation pipeline with a single general-purpose, learned model that connects
low-level perception with physical prediction.

\begin{figure}
\setlength{\unitlength}{0.5\columnwidth}
\begin{picture}(1.0,1.25) \linethickness{0.5pt}
    \put(0.0,-0.05){\includegraphics[width=1.0\columnwidth]{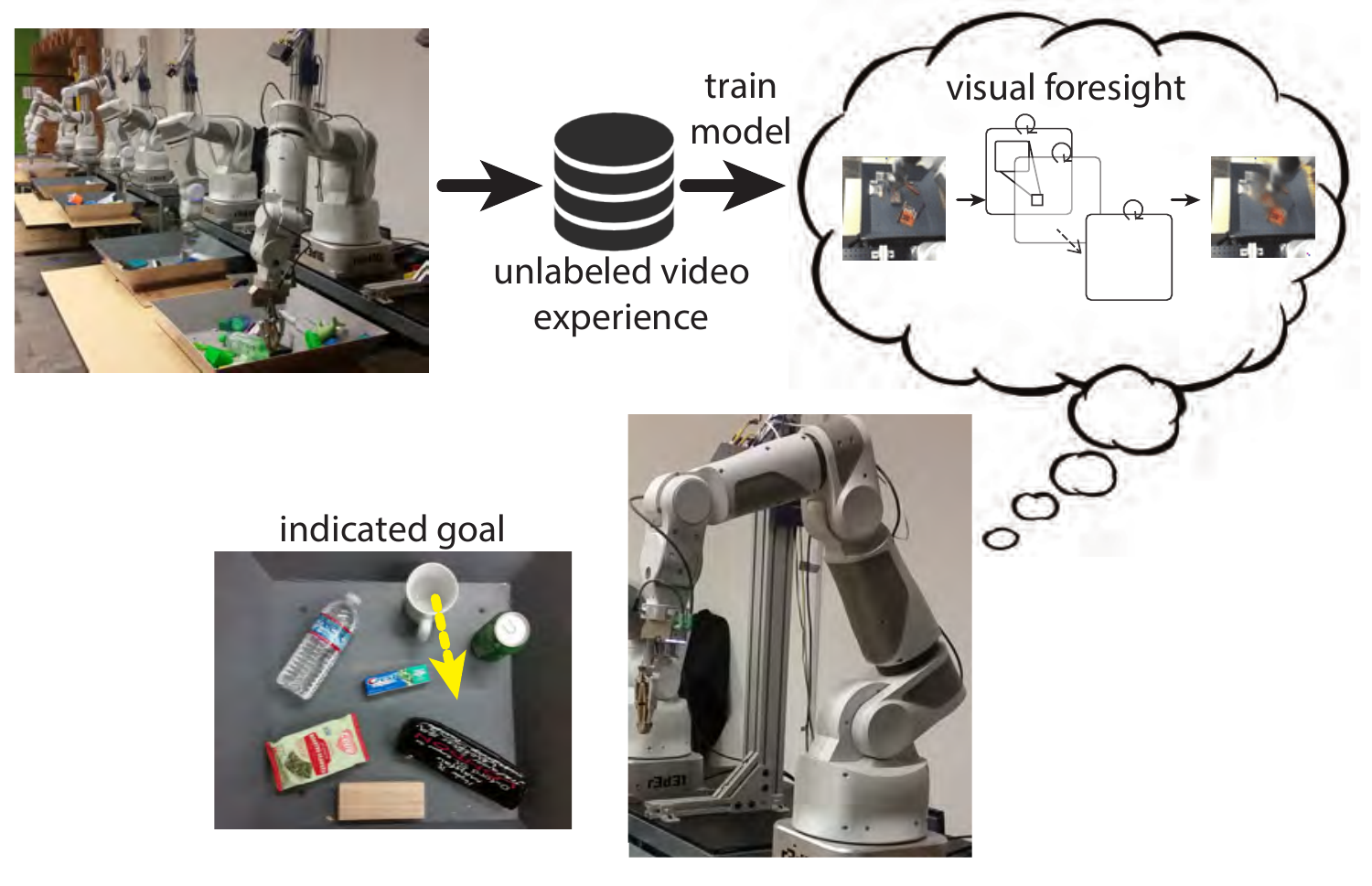}}
\end{picture}
\caption{Using our approach, a robot uses a learned predictive model of images, i.e. a visual imagination, to push objects to desired locations.
    \label{fig:teaser}
}
\end{figure}

In this paper, we take a small step in the direction of this goal, by demonstrating an approach for combining a learned predictive model of raw sensory observations with model-predictive control (MPC).
Unlike most methods for robotic learning, our approach requires minimal human involvement and can learn in an entirely self-supervised fashion, without a detailed reward function,
an image of the goal, or ground truth object pose information.
At test time, we define the task objective as moving a pixel or a group of pixels from their current position to a desired goal position (e.g. see Figure~\ref{fig:teaser}).
This goal description allows us to specify how the robot
should affect objects in its environment. Also at test time, our method optimizes for the sequence of actions that will move the pixels as desired.
The actions are continuously replanned as the robot executes the task and receives new observations, allowing the method to correct for mispredictions.

%We define the task objective as
%moving a pixel or group of pixels from their current position to a desired goal position. This goal specification allows us to specify how the robot should affect objects in its environment, without
%requiring an image of the goal or ground truth pose information. At the beginning of an episode,
%we optimize for a sequence of actions to move the pixels as desired, using a stochastic optimization algorithm.
%After the robot starts execution, we reactively re-plan actions
%using stochastic search, enabling the algorithm to incorporate the latest observations and correct for mis-predictions.

% TODO - reference this figure in the text.

The primary contribution of our paper is to demonstrate that deep predictive models of video can be used by real physical robotic systems to manipulate previously unseen objects. To that end, we present an MPC algorithm based
on probabilistic inference through a learned predictive image model that allows a robot to plan for actions that move user-specified objects in the environment to user-defined locations.
We apply a deep predictive model demonstrated on video prediction in prior work \cite{ulpi-fgl-16}, and show how the ability of this model to learn implicit stochastic pixel flow can be leveraged within a probabilistic MPC framework.
To evaluate the feasibility of robotic manipulation with learned video prediction models, we restrict the problem setting to nonprehensile pushing tasks.
By planning through the model in real time, we show that it is possible to learn nonprehensile manipulation skills with fast visual feedback control in an entirely self-supervised setting, using only unlabeled training
data and without camera calibration, 3D models, depth observations, or a physics simulator. Although the pushing tasks are relatively simple, the performance of our method suggests
that the underlying predictive model learns a sufficiently detailed physical model of the world for basic manipulation.
To the best of our knowledge, our work is the first instance of robotic manipulation using learned predictive video models with generalization to new, previously unseen objects.

%This replanning is done five times per second, allowing for fluid control. % Roberto said to take this out
%In our prior work, we developed the dynamic neural advection model for video prediction~\cite{ulpi-fgl-16},
%which internally predicts optical flow to construct the next image from the current image, rather than generating image pixels from the internal model state.
%As a result, the model can reuse object appearance information from previous frames, and both the model and the MPC controller can adapt to previously unseen, novel objects.
%The deep predictive model, described further in Section~\ref{sec:background}, is not a contribution of this work. Our primary contribution is a method, detailed in Section~\ref{sec:method}, for combining such
%a predictive model of images with MPC for performing vision-based, nonprehensile manipulation tasks.  % Roberto said that this was unclear. Now it should be clear.
%\todo{mention baselines and how it compares}

%Things to include
%- modest requirements of robot capabilities (don't need super accurate encoders or actuation -- soft robotics, cheap robots)
%- very few assumptions about the amount of human supervision (don't need to have the environment reset, don't need a reward function persay)
%- humans and animals don't have super precise proprioception, thus this biologically-inspired solution might be more scalable in the long run,
%neuroscience reference (probably in the intro, not here), efference copy? look at poking paper/ask Pulkit or Mayur

%%%%%%%%%%%%%%%%%%%%%%%%%%%%%%%%%%%%%%%%%%%%%%%%
\section{Related Work}
\label{related}

Standard model-based methods for robotic manipulation might involve estimating the physical properties of the environment, and then solving for the controls based on the known laws of physics
~\cite{osf-k-87,mirm-mls-94,pfnm-ds-12}. %~\cite{osf-k-87,mirm-mls-94,mrm-m-01}.
This approach has been applied to a range of problems, including robotic pushing~\cite{mpmp-m-86,vbp-smos-93,ppop-ches-11}. Despite the extensive work in this area, tasks like pushing an unknown object to a desired position
remain a challenging robotic task, largely due to the difficulties in estimating and modeling the physical world~\cite{mmwp-ybfr-16}. Learning and optimization-based methods have been applied to various parts of the
state-estimation problem, such as object recognition~\cite{vlr3d-mn-95}, pose registration~\cite{orfpr-cbsf-09}, and dynamics learning~\cite{ldd-etb-13}.
However, estimating and simulating all of the details of the physical environment is exceedingly difficult, particularly for previously unseen objects,
and is arguably unnecessary if the end goal is only to find the desired controls.
For example, simple rules for adjusting motion, such as increasing force when an object is not moving fast enough, or the gaze heuristic~\cite{pp-mrd-03}, can be used to
robustly perform visuomotor control without an overcomplete representation of the physical world and complex simulation calculations.
Our work represents an early step toward using learning to avoid the detailed and complex modeling associated with the fully model-based approach.

Several works have used deep neural networks to process images and represent policies for robotic control, initially for driving tasks~\cite{alvinn-p-89,llrv-hsbes-09}, later for robotic soccer~\cite{rghl-rlrs-09},
and most recently for robotic grasping~\cite{sss-pg-16,lhecrg-lpkq-16} and manipulation~\cite{eetdvp-lfda-16}. Although these model-free methods can learn highly specialized and proficient behaviors, they recover a task-specific policy rather than a flexible
model that can be applied to a wide variety of different tasks. The high dimensionality of image observations presents a substantial challenge to model-based approaches, which have been most successful
for low-dimensional non-visual tasks~\cite{dr-pilco-11} such as helicopter control~\cite{rlhf-acqn-07}, locomotion~\cite{tet-sscb-12}, and robotic cutting~\cite{dmpc-lks-15}.
Nevertheless, some works have considered modeling high-dimensional images for object interaction.
For example, Boots et al.~\cite{bbf-lpmdc-14} learn a predictive model of RGB-D images of a robot arm moving in free space. Byravan et al.~\cite{se3-bf-16} learn a model similar to the one that we use here based on prior work~\cite{ulpi-fgl-16},
except that it predicts 3D rigid motions for constructing future depth images instead of pixel flow transformations for RGB images. These three works propose video prediction models, but do not demonstrate a method
for using the learned model for control.

Agarwal et al.~\cite{ltpbp-anaml-16} learn an inverse model that can be
used for poking objects, but do not demonstrate generalization to new objects, nor planning capabilities.
Other works have used model-based approaches with a learned low-dimensional embedding of images~\cite{bsg-psr-11,lrv-arl-12,e2c-wsbr-15,ftddla-dsae-15}, defining the objective using an
image of the goal; however, such methods have been demonstrated either on simple synthetic images, or without generalization to new objects.
Our approach does not require an image of the goal and demonstrates generalization to new objects.
%%SL.9.10: maybe elaborate a little bit on these points -- explain why not requiring an image of the goal is good and why we can do it (because our model explicitly models pixel motion), and why generalization to new objects is important and why prior works have trouble with it (this is basically the large self-supervised dataset argument)

Our approach is related to visual servoing, which performs feedback control on features in an image~\cite{ecr-navsr-92,mkd-vbcqp-14,whb-reecu-96} or the image pixels themselves~\cite{bbm-cm-15}.
Unlike visual servoing methods, our approach performs feedback control at the pixel level using a model learned entirely from unlabeled, real-world video data, without requiring any explicit camera calibration.
As demonstrated in our experiments, our method can be used to move image pixels that are not directly actuated by the robot's joints -- that is, perform nonprehensile manipulation.
This is generally outside of the capability of standard model-based visual servoing techniques, due to the inherent discontinuities and complex, unknown dynamics.
%Our method uses a video prediction model to plan controls. Predictive models of images have also been used for exploration in simulated reinforcement learning
%problems~\cite{acvpa-oglls-15}.
%%SL.9.10: could probably omit this

%Drawbacks (maybe to mention elsewhere): requires a lot of data and compute power, have to predict entire images, no modular concept of objects for long-term planning

%%%%%%%%%%%%%%%%%%%%%%%%%%%%%%%%%%%%%%%%%%%%%%%%
\section{Background}
\label{sec:background}

\begin{figure*}[t]
\centering
\includegraphics[width=\textwidth]{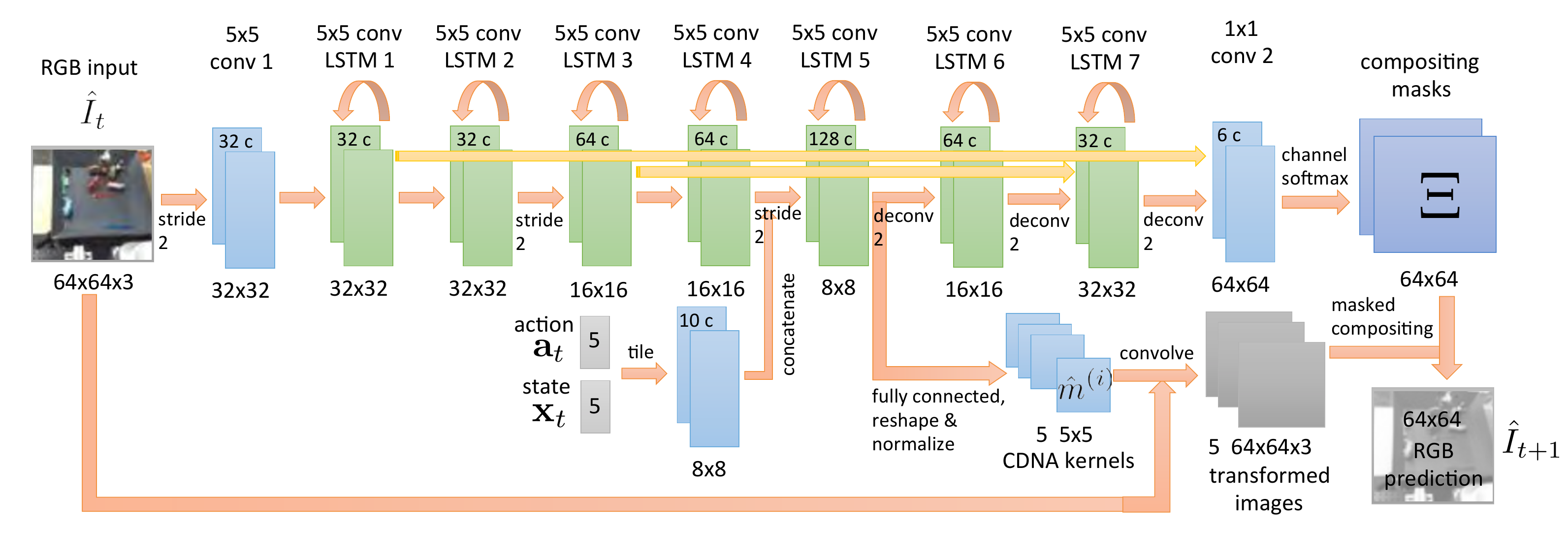}
\vspace{-1cm}
    \caption{Our video prediction model predicts stochastic pixel flow transformation from the current frame to the next frame, which allows it to generate predictions for subsequent images conditioned on a sequence of future actions. Predicted stochastic flow
        is parameterized by a set of normalized convolution filters that give rise to an independent Gaussian distribution over future images. Internally, our model predicts multiple stochastic flow channels, in order to process multiple separate moving objects.
        These channels are composited using learned object masks. The only supervision to the model consists of video, action, and state sequences, with no explicit supervision over flows or masks. The flows and masks therefore are an emergent property of the model, which we will exploit to perform visual MPC using high-level user commands.
        Further details about the model may be found in prior work~\cite{ulpi-fgl-16}.
    }
    \label{fig:model}
\end{figure*}

To perform visual model-predictive control, we use a deep predictive model of images trained on a large dataset of robotic pushing experience. In this section, we introduce notation and briefly describe
the data collection process and model. Our dataset includes $50,000$ pushing attempts collected using $10$ 7-DoF arms, involving hundreds of
objects.\footnote{We have made the entire dataset publicly available for download at \hspace{0.1in} \url{http://sites.google.com/site/brainrobotdata}}
During data collection, a bin containing between $10$ and $20$ objects is positioned in front of each robot, as shown in Figure~\ref{fig:teaser}.
Each push attempt starts with the robot randomly
selecting an initial pose along the outside border of the bin. Subsequently, the robot iteratively selects a new target gripper pose $\action_t$ and moves towards it.
A new target pose is chosen at each time step $t\in\{1,\dots,T\}$, where $T$ is the episode length.
The gripper pose is represented in the coordinate frame of the robot base as the positional coordinate in $3D$ and rotations pitch and yaw, while roll is kept constant. Thus, $\action_t$ is a vector of length 5.
During data collection, the commanded poses are chosen randomly. The robot records sensor readings at 10~Hz, including the current gripper pose $\state_t$, the commanded gripper pose $\action_t$,
and the RGB camera image of the workspace, $\image$, and selects new commanded poses at 5~Hz.

The inputs to the predictive model $\model$ at each timestep consist of the current and previous images, which we will denote $I_{0:1}$ and the current and previous end-effector poses $\state_{0:1}$,
as well as a sequence of future commands $\action_{1:H_p}$, where $H_p$ is the prediction horizon used during training.
Pairs of images and states are fed into the model to allow it to estimate the current velocities of moving objects.
The model is trained to predict a distribution over the sequence of future image frames $I_{2:H_p+1}$ that result from executing the actions $\action_{1:H_p}$.
However, since the full posterior over images is extremely high dimensional, we use a relatively simple factorization where each pixel is drawn from an independent Gaussian distribution.
Although this assumption is simplistic, it is a common simplification in video and image generation models~\cite{vpbmse-mcl-16,aevb-kw-13}.
The model is trained with maximum likelihood, which results in a mean squared error objective. We can therefore express the model probabilistically as $p_\model(I_{2:H_p} | I_{0:1}, \state_{0:1}, \action_{1:H_p})$.

As shown in Figure~\ref{fig:model}, the model uses a convolutional LSTM to predict the images. As an intermediate step, the model outputs a probabilistic flow map at each time step $t$,
which we denote $\flow_t$, that describes a linear transition operator that can be applied to each pixel in the preceding image. Intuitively, we expect different objects in different regions of the image
to be moving in different ways. The model uses a set of normalized convolution kernels $\{m_c\}$ to capture the motion of object $c$ by providing a distribution over nearby pixel locations at the next timestep.
These object motion predictions are combined using masks $\{ \Xi_c \}$ which specify the positions of the objects.
The flow map $\flow_t(x,y,k,l)$ denotes the probability of pixel $(k,l)$ at time $t+1$ originating from location $(x,y)$ at time $t$, and is given by the following.
\[
    \flow_t(x,y,k,l) = \sum_c  \Xi_{c,t}(x,y) \hat{m}_{c,t}(k,l)
\]
The model can then output a prediction for the mean of the distribution over the next image, which we denote $\hat{I}_{t+1}$, according to the following equation:
\[
    \hat{I}_{t+1}(x,y) =  \sum_{k \in (-\kappa,\kappa)} \sum_{l \in (-\kappa, \kappa)} \flow_{t}(x,y,k,l) \hat{I}_{t}(x-k,y-l)
    %\sum_c \Xi_{c,t} \circ [ \sum_{k \in (-\kappa,\kappa)} \sum_{l \in (-\kappa, \kappa)} \hat{m}_{c,t}(k,l) \hat{I}_{t}(x-k,y-l)]
\]
Note that this equation is an untied convolution operation between the filters $\flow_t$ and the image $\hat{I}_t$, since the filters are different for each pixel location $(x,y)$.
The operator $\flow_t$ can also be seen as the transition operator in a Markov chain.
We will use $\hat{I}_{t+1} = \flow_t \odot I_t$ to denote the application of the flow operator. Since this operator is linear,
the distribution over a subsequent image can be obtained simply by transforming the mean of the distribution over the current image.
The predicted image mean $\hat{I}_{t+1}$ is fed back into the network recursively to generate the next flow and image in the sequence.
Note that no explicit flow supervision is provided to the network: training uses only the raw image pixels for supervision.
The stochastic flow maps are an implicit, emergent property of the model, but one that will prove useful in visual MPC, as we will discuss in the following section.
For convenience, we will use $\model(I_{t-1:t}, \state_{t-1:t}, \action_{t:H}) = \flow_{t:H}$ to denote the function that uses the learned model to output a sequence of flows conditioned on pairs of images and states, as well as a sequence of future actions, for some horizon $H$. We can therefore define the predicted image distribution as
\begin{align*}
    p_\model(I_{t+1} | I_{t-1:t}, \state_{t-1:t}, \action_{t}) &= \mathcal{N}(\flow_t \odot I_t, \sigma^2\mathbf{I}) \\
&= \mathcal{N}( \model(I_{t-1:t}, \state_{t-1:t}, \action_{t}) \odot I_t, \sigma^2\mathbf{I}),
\end{align*}
\noindent where $\sigma^2\mathbf{I}$ is a constant diagonal covariance. Predictions for subsequent images $I_{t+k}$ can then be made recursively, with mean given by $\model(I_{t-1:t}, \state_{t-1:t}, \action_{t:t+k-1}) \circ \hat{I}_{t+k-1}$.

We use the same network architecture as in prior work~\cite{ulpi-fgl-16}, except for the addition of
layer normalization~\cite{ln-bkh-16} after each layer for more robust training, and the use of $5$ masks instead of $10$ for slightly faster computation.

%For more information about the model, see Finn et al.~\cite{ulpi-fgl-16}.
%This transformation can be parameterized using convolutions followed by masking, hence it is called convolutional dynamic neural advection.

%Each type of motion lasted for approximately 3-5 seconds.
%Between each motion, the arm was programmed to move out of the camera scene, and an image was recorded.

%%%%%%%%%%%%%%%%%%%%%%%%%%%%%%%%%%%%%%%%%%%%%%%%
\section{Visual MPC with Learned Video Prediction Models}  % Or just Visual Model Predictive Control
\label{sec:method}

At test time, our system receives a high level goal from the user, which we discuss in Section~\ref{sec:goal}, and then uses model-predictive control (MPC) together with our deep video prediction model to choose controls that
will realize the user's commanded goal, as shown in Figure~\ref{fig:teaser}. Because video prediction models do not explicitly model objects, we cannot represent the task goal using standard notions of object poses.
In this section, we outline exactly how the objective is specified and represented, and how to probabilistically evaluate candidate actions using the
model and goal representation. Lastly, in Section~\ref{sec:planning}, we present our method for planning and replanning actions using MPC, which involves rolling out the video prediction model $H$ timesteps into the future.
We show the full algorithm in Algorithm~\ref{alg:vmpc}, which combines the learned model, the high-level goal specification, and MPC to probabilistically plan future actions.

\subsection{Specifying Goals with Pixel Motion}
\label{sec:goal}

To plan with a predictive model of images, we need a way to specify task objectives that can be automatically evaluated for each of the model's predicted visual futures.
In this work, the goal is specified by the user in terms of pixel motion. Intuitively, the user specifies a pixel in the image and tells the robot where that pixel should be moved to.
The user selects one or more desired source pixels in the initial image, which we will denote $\pixel_0^{(1)},\dots,\pixel_0^{(P)}$, with each pixel's position given by $(x_d^{(i)}, y_d^{(i)})$,
and then specifies a corresponding goal position for each of those pixels, which we will denote $\goal^{(1)},\dots,\goal^{(P)}$, with each goal position given by $(x_g^{(i)},y_g^{(i)})$.
With this goal specification, the robot can plan to move the objects for which the selected pixels belong.
This kind of goal specification is quite general and can be used to command arbitrary rearrangements of objects, such as clearing a table.
In a practical application, the commands themselves could be issued either directly by a human user selecting points in an image, or by a higher-level planning process.
This goal representation is easy to specify, does not require instrumentation of the environment e.g. via motion capture or AR markers, and can
represent a variety of object manipulation tasks. Unlike most approaches to robotic manipulation, it also does not use or need an explicit representation of objects.
Example pixel motion goals are shown in Figure~\ref{fig:qual}.
%%SL.9.15: it's pretty distracting to forward-reference a figure that is so much later in the paper.
%%SL.9.15: can we replace the image I with \mathbf{I} to emphasize that it is a 2D array?

\subsection{Evaluating Actions with Implicit Pixel Advection}
\label{sec:inference}

Here, we describe how our method evaluates a proposed action sequence using probabilistic inference under the learned predictive model $\model$. For simplicity, first consider the goal of moving a single designated pixel $\pixel_0 = (x_d,y_d)$
in the initial image to a goal position $\goal = (x_g, y_g)$. We need to evaluate the probability of achieving this goal under the model $\model$ for a given initial image $I_t$, intial state $\state_t$, and sequence
of $H$ future actions $\action_t,\dots,\action_{t+H-1}$.
As described in Section~\ref{sec:background}, our video prediction model predicts stochastic flow operators $\flow_{t+k}$ at each time step, which can be used to transform prior images or independent
Gaussian pixel distributions into independent Gaussian pixel distributions for the next image, according to $\hat{I}_{t+k} = \flow_{t+k} \odot \hat{I}_{t+k-1}$.
We can also use these stochastic flow operators to predict how individual pixels will move, conditioned on a sequence of actions. Since the flow operators are stochastic,
they provide a distribution over pixel motion conditioned on the actions. This suggests a natural approach to planning using maximum likelihood: determine the sequence of actions that maximizes the
probability of the designated pixel moving to the goal position.

%We now outline this process in detail.
Formally, we construct an initial probability distribution over the designated pixel's position at the current time step $t$, which we denote $P(\pixelstate_t)$.
We assume that the current position of the designated pixel $\pixel_t = (x_d,y_d)$ is known at time $t$, as it is provided by a user at time $t=0$ can be tracked thereafter. Thus, we define the distribution at the current
time step $t$ as
\[
P(\pixelstate_t) = \left\{ \begin{array}{l} 1 \text{ if } \pixelstate_t = \pixel_t \\ 0 \text{ otherwise} \end{array} \right.
\]
Our goal is to compute the distribution over the designated pixel's position at the end of the MPC horizon, denoted $P(\pixelstate_H | I_{t-1:t}, \state_{t-1:t}, \action_{t:t+H-1}, \pixel_t)$,
and choose the actions that maximize the probability
$P(\pixelstate_H = \goal | I_{t-1:t}, \state_{t-1:t}, \action_{t:t+H-1}, \pixel_t)$. This corresponds to probabilistic inference over the action sequence $\action_{t:t+H-1}$.
Recall that our predictive model outputs a stochastic flow operator $\flow_t$ at each time step. The stochastic flow corresponds to a stochastic transition operator over the pixel positions,
and we can therefore propagate the distribution over the designated pixel forward in time using $\flow_t$. Let $\mathbf{P}_{t+k}$ denote the 2D conditional distribution $P(\pixelstate_{t+k} | I_{t-1:t}, \state_{t-1:t},
\action_{t:t+k}, \pixel_t)$, we then have:
\begin{align}
    \begin{split}
        \label{eq:eval}
\mathbf{P}_{t+k+1} &= \flow_{t+k} \odot \mathbf{P}_{t+k} \\
&= \model(I_{t-1:t}, \state_{t-1:t}, \action_{t:t+k}) \odot \mathbf{P}_{t+k}.
    \end{split}
\end{align}
Applying this computation recursively for $H$ steps, we obtain $\mathbf{P}_{t+H-1} = P(\pixelstate_{t+H-1} | I_{t-1:t}, \state_{t-1:t}, \action_{t:t+H-1}, \pixel_t)$, which describes the probability distribution over the position
of the designated pixel at the end of the MPC horizon. The probability of successfully moving the designated pixel to the goal location is then given simply by $\mathbf{P}_{t+H-1}(\goal)$.
This process can easily be extended to multiple pixels by computing the probability
of success for each pixel separately, and summing the log-probabilities.

Interestingly, this process makes use of the model's implicit flow predictions, instead of the predicted images themselves. However, the ability of the model to predict images is what enables
us to train it without explicit flow supervision.

\subsection{Sampling-Based Model-Predictive Control with Deep Models}
\label{sec:planning}

\begin{algorithm}[t]
{\small
    \caption{Visual MPC with Deep Predictive Models}
\label{alg:vmpc}
\begin{algorithmic}[1]
%\item \textbf{inputs:} predictive model $\model$, designated pixels $\{\pixel^{(k)}_0 \}$, pixel goal positions $\{\goal^{(k)} \}$
\item \textbf{inputs:} predictive model $\model$, designated pixel $\{\pixel_0 \}$, pixel goal position $\{\goal \}$
    \FOR{$t=1...T$}
        \STATE Initialize $Q_1$ with uniform distribution.

        \FOR{$j=1...J_t$}
            \STATE Sample $M$ action sequences $\{\action^{(m)}_{t:t+H-1}\}$ from $Q_j$.
            \STATE Use model $\model$ to compute the distributions over future pixel locations $\mathbf{P}^{(m)}_{t+H-1}$ using Equation~\ref{eq:eval}
            %\STATE Determine set of $K$ best samples with highest probability of success $\mathbf{P}^{(m)}_{t+H}(\goal)$  % TODO - make this work for multiple pixels
            \STATE Fit multivariate Gaussian distribution $Q_{j+1}$ to $K$ samples with highest probability of success $\mathbf{P}^{(m)}_{t+H-1}$
        \ENDFOR

        \STATE Execute action $\action^*_t$ with highest probability of success
        \STATE Observe new image $I_{t+1}$.
        \STATE Set next designated pixel location $\pixel_{t+1}$ using optical flow computed from image observations $I_{t:t+1}$, and $\pixel_{t}$.

    \ENDFOR
\end{algorithmic}
}
\end{algorithm}

The goal of our MPC-based controller is to determine the sequence of actions that maximizes the probability of the designated pixel or pixels being moved to their corresponding goal locations.
The previous section describes how we can evaluate a candidate action sequence $\action_{t:t+H-1}$ to determine the probability that it will move the designated pixel to the goal location,
given by $\mathbf{P}_{t+H-1}(\goal) = P(\pixelstate_{t+H-1} = \goal | I_{t-1:t}, \state_{t-1:t}, \action_{t:t+H-1}, \pixel_t)$. In order to use this evaluation to choose the best actions,
we perform an optimization over a short horizon of actions at each time step, using a stochastic optimization algorithm called the cross-entropy method (CEM)~\cite{cem-rk-13}. This procedure
is outlined in Algorithm~\ref{alg:vmpc}, and described next.

At each time step $t$, we sample $M$ action sequences of length $H$, $\{ \action_t^{(m)},\dots,\action_{t+H-1}^{(m)}\}$, and compute the probabilities of success for each one, denoted by
\mbox{$\mathbf{P}^{(m)}_{t+H-1}(\goal) = P(\pixelstate_{t+H-1} = \goal | I_{t-1:t}, \state_{t-1:t}, \action_{t:t+H-1}^{(m)}, \pixel_t)$}.
We then select the $K$ action sequences with the highest values of $\mathbf{P}^{(m)}_{t+H-1}(\goal)$, fit a multivariate Gaussian distribution to these $K$ selected
action sequence, and resample a new set of $M$ action sequences from this distribution.
The new set of action sequences improves on the previous set, and the resampling and refitting process is repeated for $J_t$ iterations. This corresponds to the CEM stochastic optimization algorithm.
At the end of the last iteration, we take the sampled action sequence $\action^*_t,\dots,\action^*_{t+H-1}$ that is most likely to be successful, and execute $\action^*_t$ on the robot.
We use $J_t=4$ iterations, $M=40$ samples per iteration, and $K=10$
samples during the initial planning phase ($t=0$) and, when replanning in real-time ($t>0$), we take $J_t=1$ iteration of $M=20$ samples,
performing just one round of sampling.
Note that each batch of $M$ samples corresponds to a forward pass through deep recurrent network with a batch size of $M$, and therefore can be parallelized very efficiently.

After the first timestep, the designated pixel locations may have changed from their initial positions due to motion of the objects.
To update the estimated position of each pixel, we compute optical flow on the previous and latest image observation, using the method
of Anderson et al.~\cite{jvrv-agbks-16}, an optical flow algorithm used in a large-scale production system. For speed, the optical flow computation is done on the CPU while evaluating the model's video predictions in parallel using a GPU.

%%%%%%%%%%%%%%%%%%%%%%%%%%%%%%%%%%%%%%%%%%%%%%%%
\section{Experiments}
\label{sec:experiments}

%\begin{figure}
%\setlength{\unitlength}{0.5\columnwidth}
%\begin{picture}(1.0,1.0) \linethickness{0.5pt}
%    \put(0.0,-0.05){\includegraphics[width=0.8\columnwidth]{robots.jpg}}
%\end{picture}
%\caption{Our method uses only unlabeled data that was collected using $10$ 7-DoF robot arms, and minimal human involvement.
%    \label{fig:robots}
%}
%\end{figure}

\begin{figure}  % maybe make this a full column
\setlength{\unitlength}{0.5\columnwidth}
\begin{picture}(1.0,1.07) \linethickness{0.5pt}
    \put(0.02,0.0){\includegraphics[width=0.29\columnwidth]{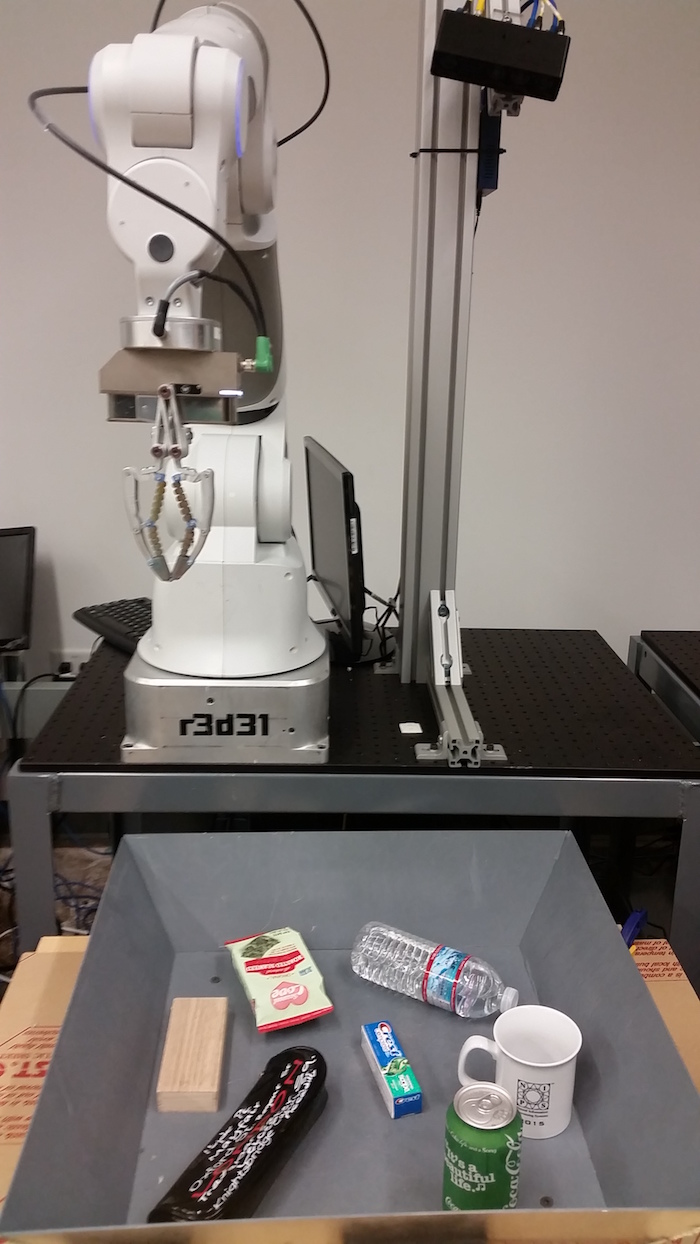}}
    \put(0.625,0.0){\includegraphics[height=0.26\columnwidth]{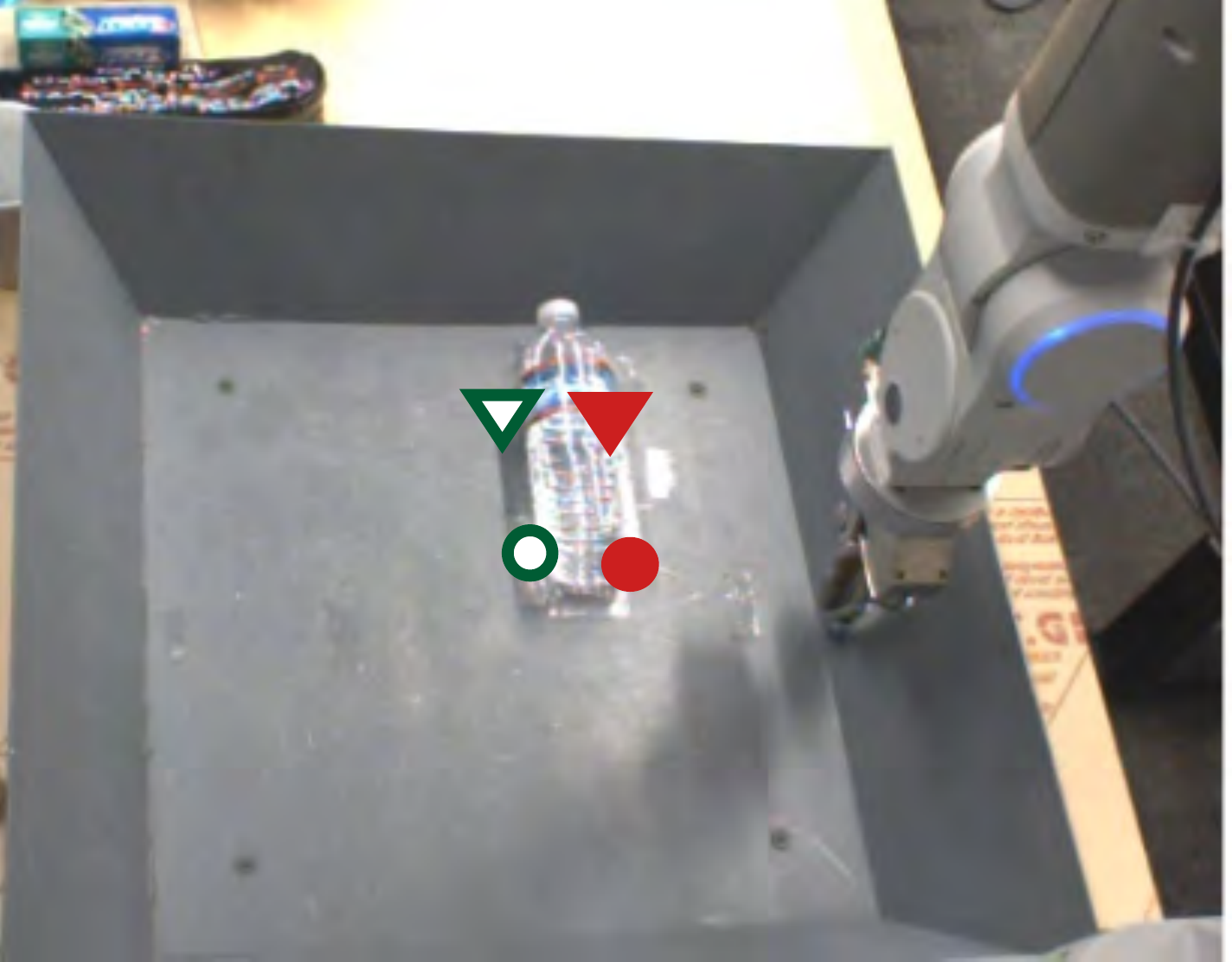}}
    \put(1.3,0.0){\includegraphics[height=0.26\columnwidth]{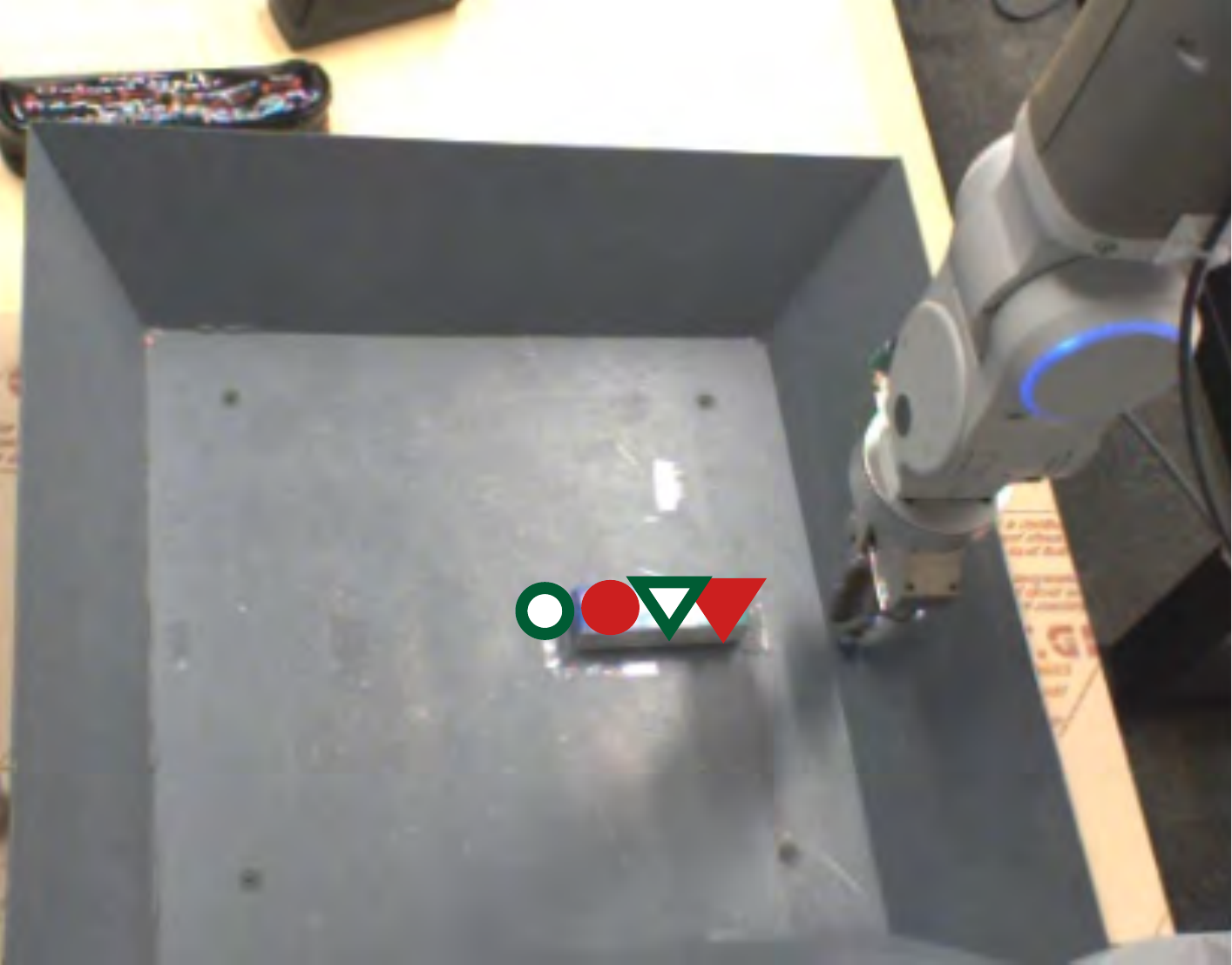}}
    \put(0.625,0.5){\includegraphics[height=0.26\columnwidth]{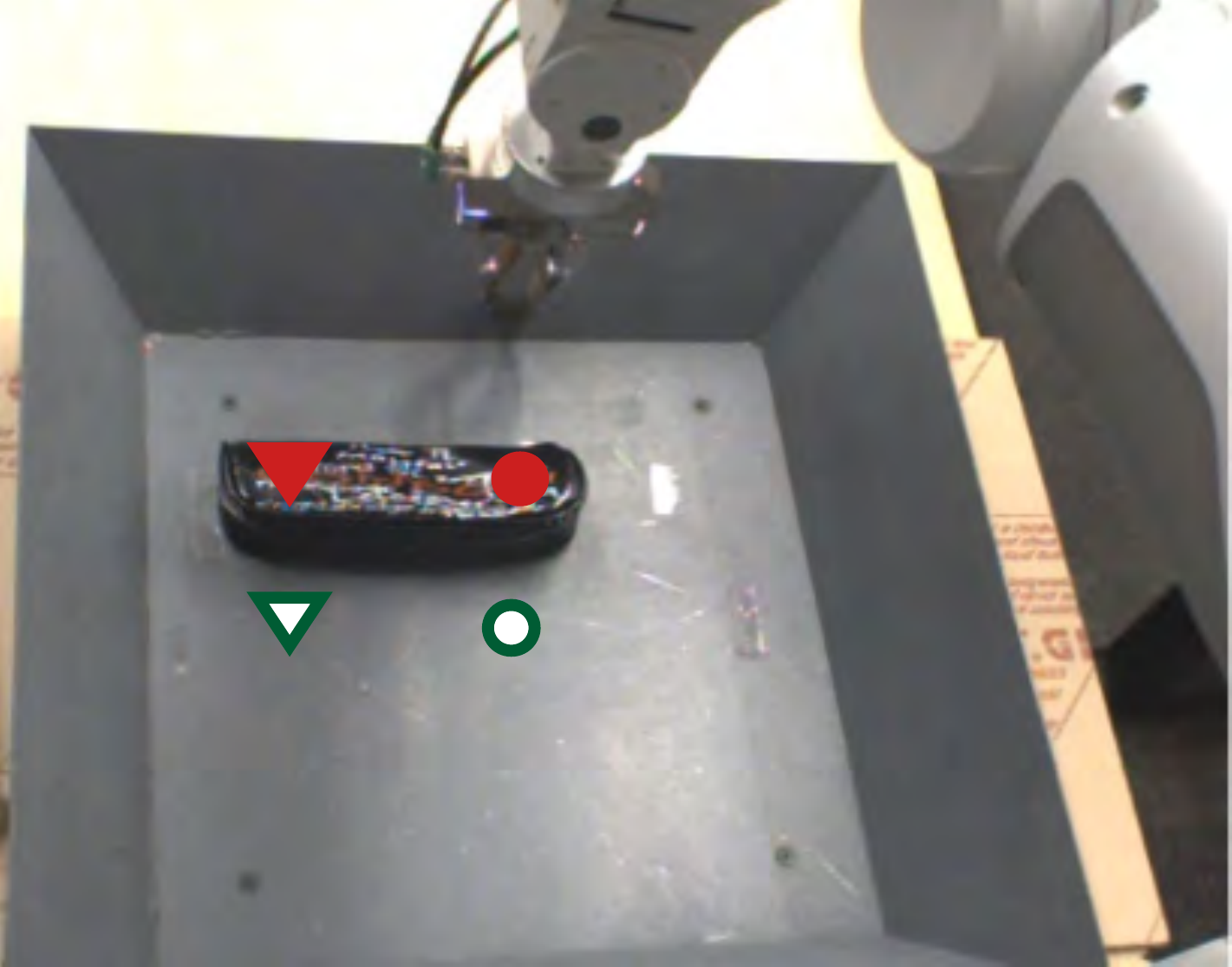}}
    \put(1.3,0.5){\includegraphics[height=0.26\columnwidth]{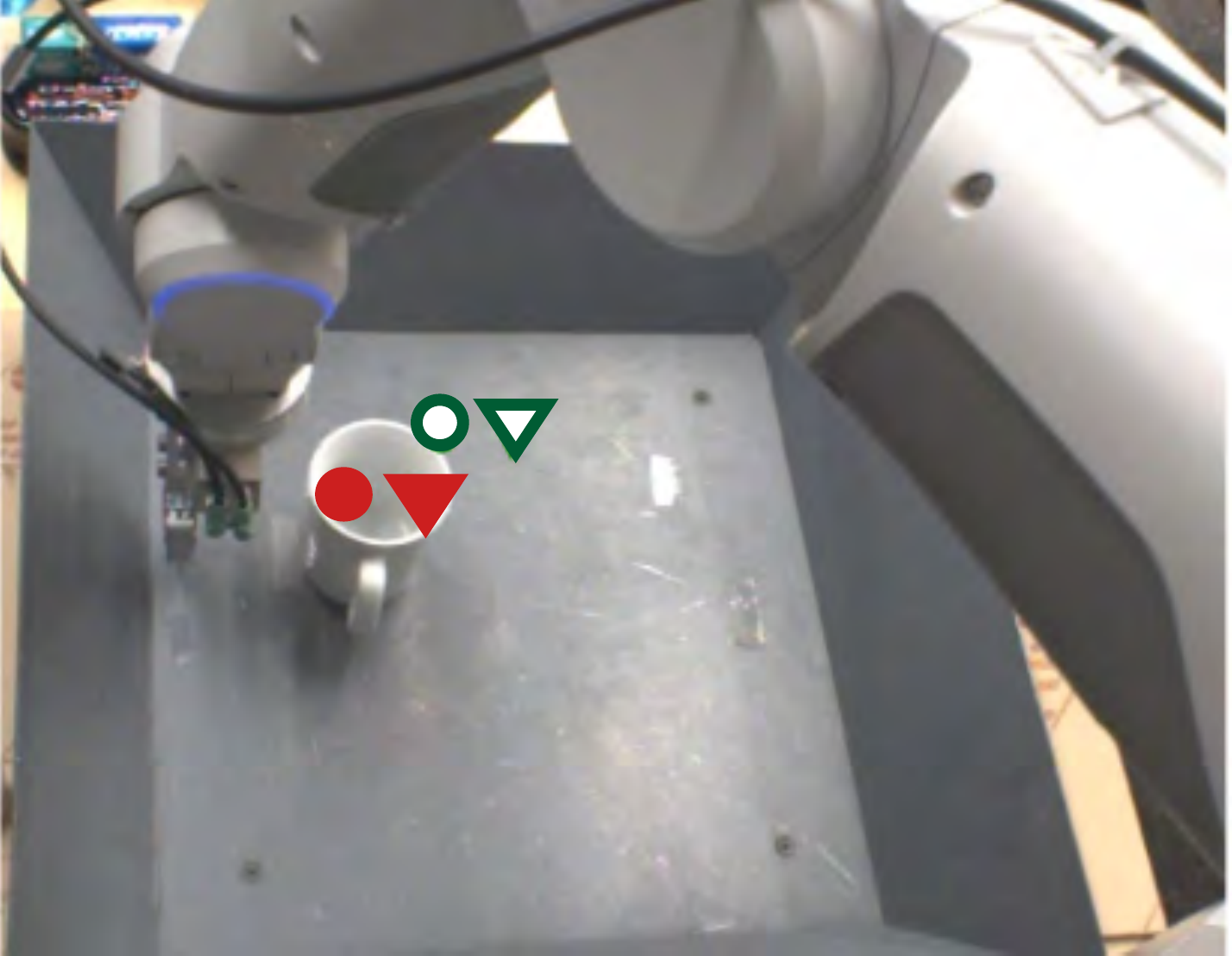}}
\end{picture}
\caption{The experimental setup is shown on the left, including the $7$ test objects previously not seen in the training set. On the right, we show four of the ten pushing tasks in the quantitative evaluation.
    The red and green-outlined markers indicate the human-specified start and goal pixel locations, respectively.
    \label{fig:robots}
}
\end{figure}

In our experimental evaluation, we aim to answer the following questions: (1) Can we use action-conditioned video prediction models to manipulate novel objects that were not previously seen during training?
(2) Can video prediction models trained entirely on raw image pixels make meaningful and nontrivial inferences about the behavior of physical objects?
To answer these questions, we conduct both qualitative and quantitative experiments, which we describe in the next sections.
We aim to answer question (1) by evaluating our method on new objects not seen during training, and we answer question (2) through comparison to baseline methods that either move the arm to user-specified positions, or use optical flow to perform continuous replanning.
We also answer question (2) through qualitative experiments aimed to construct physically nuanced pushing scenarios that require reasoning about rotations and centers of mass.

% TODO - talk about scalability
Note that the intent of our experiments is not to demonstrate that our approach provides the highest accuracy or performance for precise nonprehensile manipulation,
but rather to demonstrate the flexibility of a data-driven, learning-based approach and illustrate that,
even with no prior knowledge about objects, physics, or contacts, a predictive model trained entirely on raw video data can still infer characteristics of the physical world that are useful for robotic manipulation.

\subsection{Experimental Setup}

We use a 7-DoF robot arm to perform the pushing tasks in our experiments, with an RGB camera positioned over the shoulder, as shown in Figure~\ref{fig:robots}.
As discussed previously, the pushing task consists of episodes of length $T=15$ where the goal is to move $M$ pixels from their current locations
$\{(x^{(i)}_s, y^{(i)}_s)\}$ to corresponding goal locations $\{(x^{(i)}_g, y^{(i)}_g)\}$. Unless otherwise specified, all objects in the experiments were not seen previously in the training set.
A video of our
experiments is available online.\footnote{See \url{https://sites.google.com/site/robotforesight/}}

%To evaluate our approach, we consider the task of nonprehensile pushing on a flat table,

We use images with a resolution of $64 \times 64$ pixels and a planning horizon of $H=3$, corresponding to about 800~ms.
This allows us to replan in real time, with new controls computed about every 200~ms.
To reduce the dimensionality of the action space and further speed up inference, we tie the commanded action across time, such that the model is considering one commanded action, kept constant for $H$ timesteps.
Because of the short time horizon, we largely consider
pushing tasks that involve fast, reactive control rather than long-term planning.
%%SL.9.14: so does that mean that in the evaluation we only plan one action? if so, that could probably be clarified further.
All of the online model computations, including replanning, are done using a standard desktop computer and a single, commercially-availble GPU.

\subsection{Quantitative Comparisons}

\begin{figure}
\setlength{\unitlength}{0.5\columnwidth}
\begin{picture}(1.0,0.85) \linethickness{0.5pt}
    \put(0.0,0.0){\includegraphics[width=0.48\columnwidth]{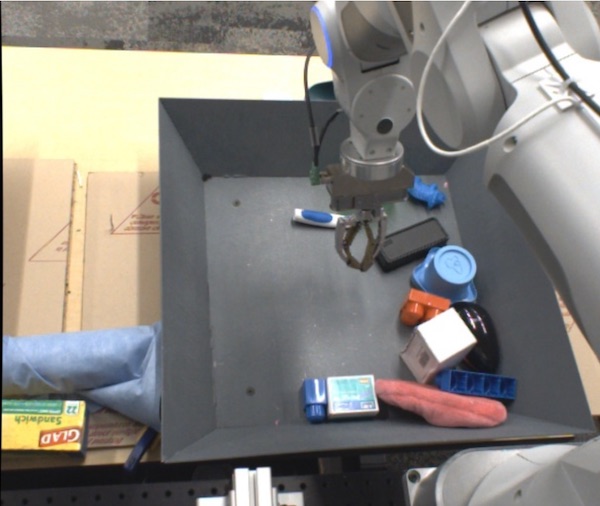}}
    \put(1.0,0.0){\includegraphics[width=0.48\columnwidth]{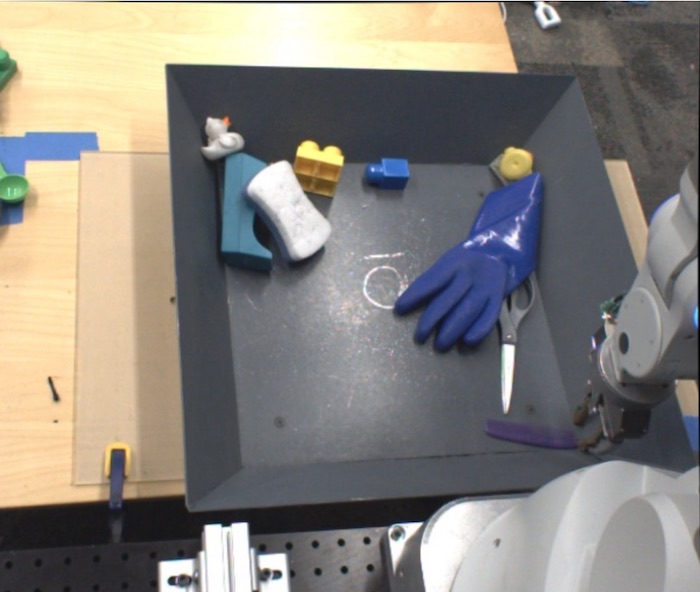}}
\end{picture}
\caption{The data for training the model was collected on $10$ robots with varying camera angles and positions. These images show the camera angles for two of the robots used during data collection. Note the difference in the position of the robot base. Due to these variations, our video prediction model learns a calibration-invariant representation of object interactions.
    \label{fig:calibration}
}
\end{figure}

We provide quantitative comparisons to three baselines, with the aim of evaluating whether or not our video prediction model has learned a meaningful and nontrivial notion of objects and physical interaction.
Recall that object identity, inertia, and contact dynamics are not provided to nor encoded in the model explicitly, but must be learned entirely from data.
Correspondingly, our baselines do not use any knowledge about the objects in the scene, but are reasonably effective for the short horizon pushing tasks that we consider.
The baselines are as follows:
\begin{enumerate}
    \item Select actions randomly from a uniform distribution.
    \item Servo the end-effector to the goal pixel position $(x_g, y_g)$, using a known camera calibration.
    \item Servo the end-effector along the vector from the current pixel position $\pixel_t = (x_t, y_t)$ to the goal $(x_g, y_g)$, with continuous replanning based on the current pixel position estimated using optical flow.
\end{enumerate}

If more than one designated pixel is specified, the last two baselines use the first pixel only.
The first baseline of randomly selecting actions serves to calibrate the difficulty of the task in choosing effective actions. The last two baselines serve as a comparison to our method to test whether our model is learning
something meaningful about physical object interaction, beyond simple motions of the arm. Note that the last two baselines require hand-to-camera calibration, which our model does not use. Since the data that was used for training our predictive model was collected on a variety of robots with varying camera placements and angles, as illustrated in Figure~\ref{fig:calibration}, our video prediction model was forced to learn a calibration-agnostic predictive strategy.
The final baseline and our method use an optical flow solver~\cite{jvrv-agbks-16} to track the position of the pixel for replanning. Note, however, that the predictive model itself does not use the optical flow to make predictions. The optical flow is used only during the replanning phase to provide the initial $P(s_t)$ distribution.
The flow solver is also used
to quantitatively evaluate the distance between the final position of the pixel and the goal position of the pixel.
%%SL.9.14: can we say something convincing about why we believe the flow solver was actually correct?

Note that we specifically choose baseline methods that do not have prior knowledge about objects or physics. It is of course possible to design a model-based pushing algorithm that uses an object detector and physics simulator
to more precisely localize and move individual objects in the scene. However, our aim is not to propose a superior method for nonprehensile manipulation, but rather to explore the capabilities and limitations of learning-based
video prediction models for performing robotic control tasks from scratch, with minimal prior knowledge.

%and more discussion/intro-style material:
%Such models are still in their infancy, with initial video prediction work providing reasonable-looking images only one or two frames into the future~\cite{beyond_mse}. However, their flexibility and minimal manual engineering requirements make them an appealing prospect for future model-based robotic control methods, and it is therefore interesting to explore their potential with the simple nonprehensile manipulations tasks that we consider in this work. As the capabilities of video prediction methods improve, we expect that prediction-based robotic control methods such as the one in this paper will improve with them, allowing for more complex skills to be executed on demand with minimal manual engineering.

Our results, shown in Table~\ref{tbl:quant}, indicate that our method is indeed able to leverage the predictive model to improve over the performance of the baselines. The performance of our method compared to the
last two baselines suggests that our method is making meaningful inferences about the motion of objects in response to the arm. Although these results leave significant room for improvement, they suggest that predictive models can be used with minimal prior knowledge to perform robotic manipulation tasks. As video prediction models continue to improve, we expect that the performance of our method will improve with them. In the next section, we analyze specific physical interactions to better understand the capabilities and limitations of our model.
%%SL.9.14: need to discuss baselines in more detail, why they do what they do, and what the results say about our method.
%%TODO(CF)

\begin{table}[h]
\caption{Quantitative comparison: mean distance between final and goal pixel positions}
\vspace{-0.4cm}
\label{tbl:quant}
\begin{center}
\begin{tabular}{|l||c|}
\hline
method & mean pixel distance \\
\hline
\hline
initial pixel position & $ 5.10 \pm 2.25$ \\
\hline
1) random actions & $4.05 \pm 1.75$ \\
\hline
2) move end-effector to goal & $3.79 \pm 2.66$ \\
\hline
\specialcell{3) move end-effector along vector\\ (with replanning)} & $3.19 \pm 1.68$ \\
\hline
visual MPC (ours) & \textbf{$\mathbf{2.52 \pm 1.06}$} \\
\hline
\end{tabular}
\end{center}
\end{table}

\subsection{Qualitative Results}

\begin{figure}  % maybe make this a full column
\setlength{\unitlength}{0.5\columnwidth}
\begin{picture}(1.0,2.0) \linethickness{0.5pt}
    \put(0.0,1.5){\includegraphics[width=1.0\columnwidth]{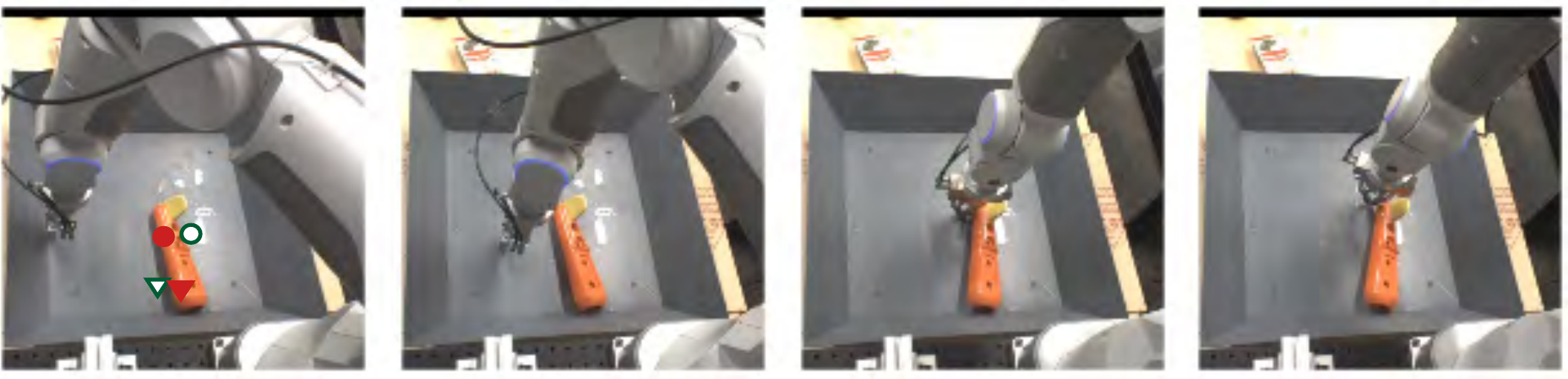}}
    \put(0.0,1.0){\includegraphics[width=1.0\columnwidth]{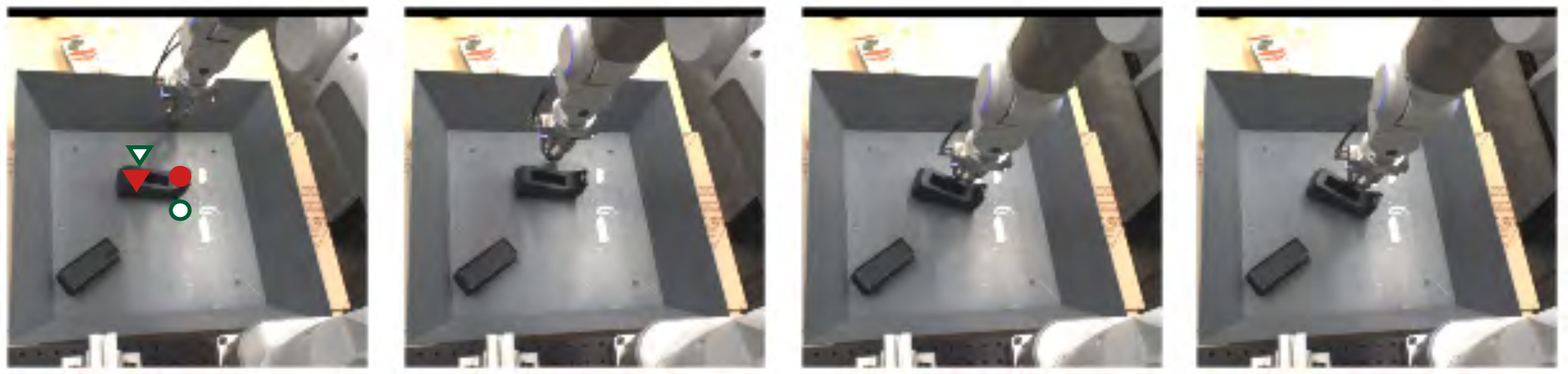}}
    \put(0.0,0.5){\includegraphics[width=1.0\columnwidth]{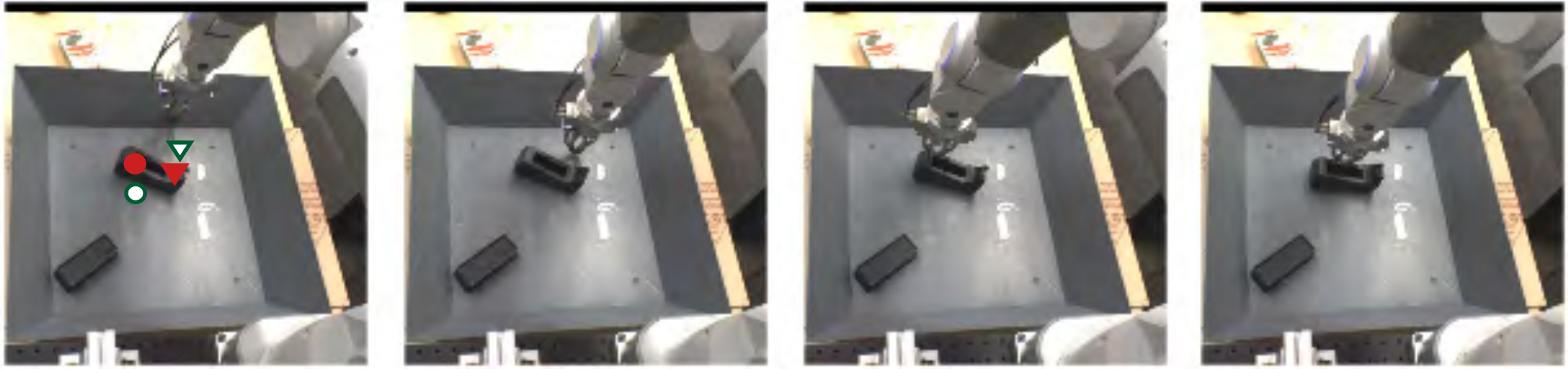}}
    \put(0.0,0.0){\includegraphics[width=1.0\columnwidth]{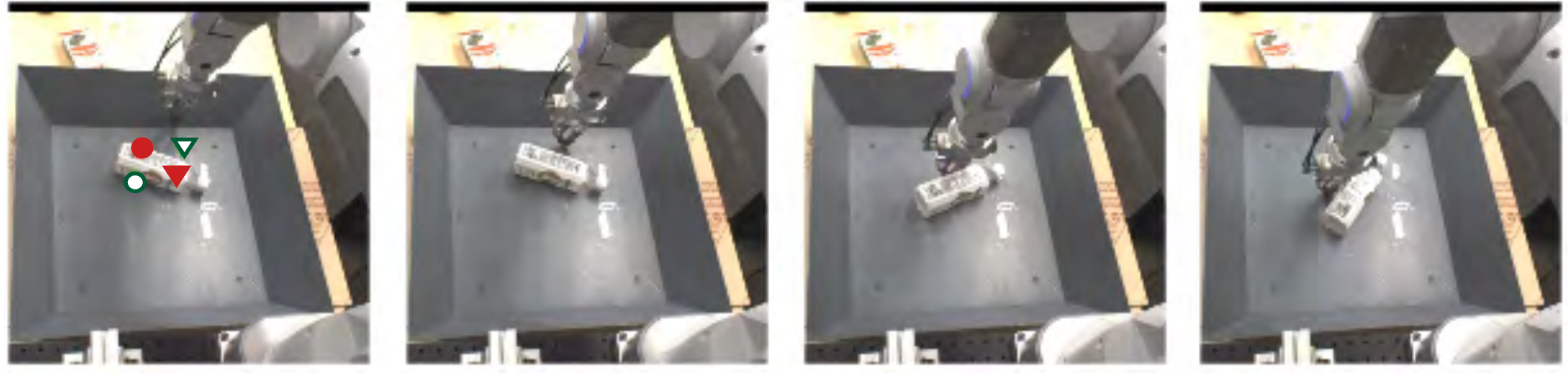}}
    \put(0.49,0.0){\line(0,1){2.0}}
\end{picture}
\caption{When commanded to rotate objects by moving their end-points in opposing directions, our method plans for actions that touch the object on one side.
    In the initial image on the left, the red markers designate the user-specified starting pixel positions, and the green-outlined markers show the corresponding goal positions.
    Note that the arm starts some distance from the object, forcing the model to plan for the right contact position to realize the rotation. The black tape in the middle two rows was in the original dataset, while
    the other two objects were not.
\label{fig:qual}
}
\end{figure}

In this section, we evaluate the capabilities and limitations of visual MPC with a set of qualitative experiments. The goal of these experiments is to determine whether or not the model can perform more complex manipulations that require reasoning about object rotations and centers of mass.
One of the benefits of planning actions with a predictive model of all pixels is that the model can be used to plan actions that affect multiple pixels,
causing them to move in different directions in a coordinated fashion. For example, object rotation can be represented by moving the pixels on opposite ends of an object in opposite directions.
To evaluate this capability, we command the robot to rotate objects by specifying opposing motions for pixels on either extreme of the object, as shown on the left in Figure~\ref{fig:qual}.
When commanded to move pixels such that an object rotates, our method is able to produce the desired rotational
motion, as seen on the right side of Figure~\ref{fig:qual}.
These examples indicate that the model can indeed be used for predicting the motion of multiple pixels and planning to affect them in a coordinated way.

%%SL.9.14: it feels like you need some kind of transition paragraph here before you get into limitations. Something to discuss the implications of the above result, and then lead into the limitations

\begin{figure}
\setlength{\unitlength}{0.5\columnwidth}
\begin{picture}(1.0,1.0) \linethickness{0.5pt}
    \put(0.0,0.5){\includegraphics[width=1.0\columnwidth]{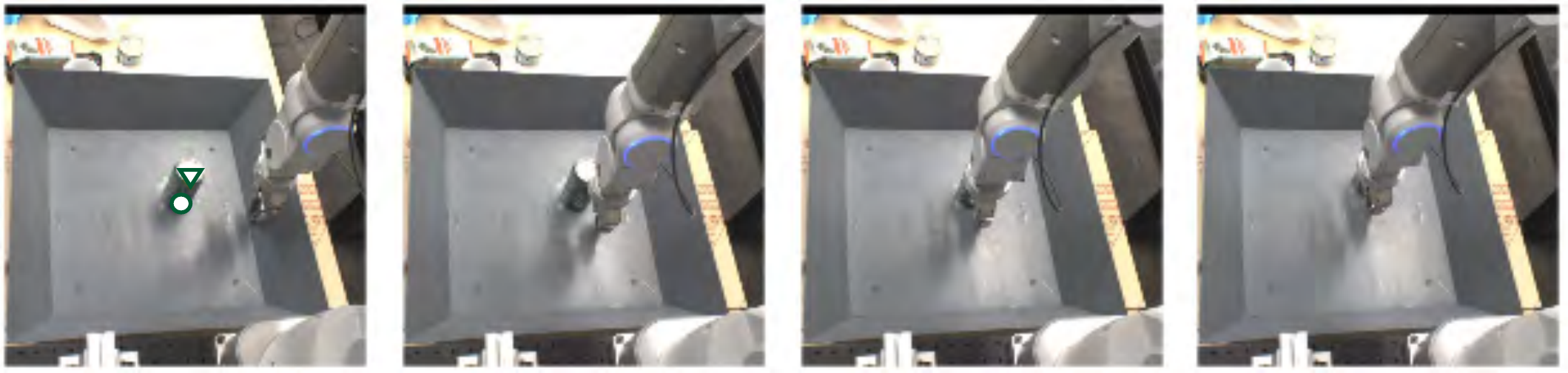}}
    \put(0.0,0.0){\includegraphics[width=1.0\columnwidth]{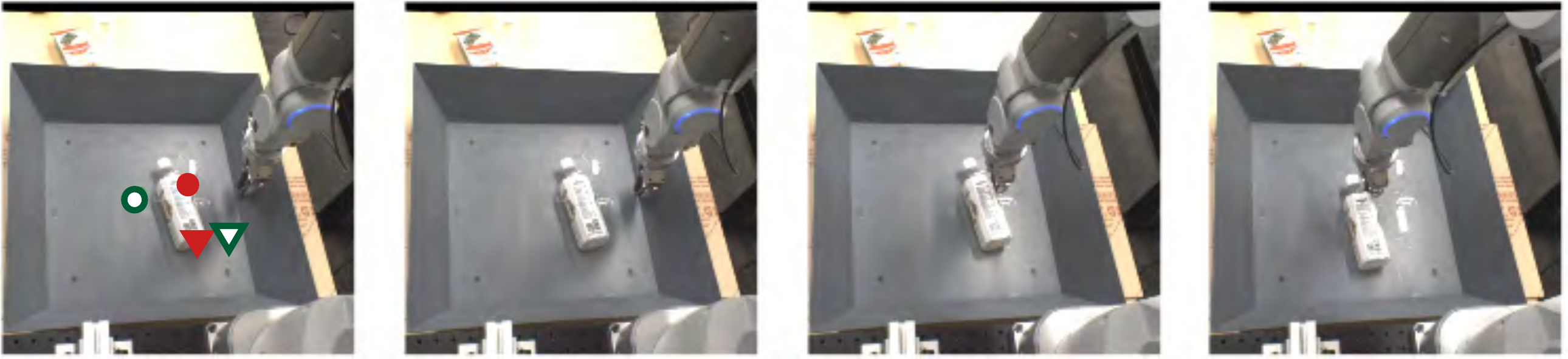}}
    \put(0.49,0.0){\line(0,1){1.0}}
\end{picture}
\caption{Two failure cases of our method. In the first scenario, the goal is to avoid moving the can. The arm moves in front of and occludes the can, occluding it, causing the estimated pixel positions to shift, and
    causing the arm to later bump into the can. In the second example,
    the goal is to rotate the bottle. Although the robot makes contact with the bottle in a reasonable location to perform the rotation, it underestimates the mass of the bottle (which is empty).
    Instead of rotating, the light bottle simply translates along with the robot's arm.
    \label{fig:failure}
}
\end{figure}

One limitation of our approach is in handling self-occlusions. Once the arm occludes the object, the designated pixel is on the arm instead of the object, and the method subsequently plans to move the arm
instead of the object into the goal position. An example of this failure case is shown in Figure~\ref{fig:failure}. Incorrect model predictions also cause failures in performance. As shown in the second row of Figure~\ref{fig:failure}, the
robot selects a sequence of actions that push the bottle too close to its center of mass, causing it to translate rather than rotate. We expect that, by improving the accuracy of video prediction models, we can reduce
such failures, and increase the general performance of the method. Video prediction is an active area of research in deep learning and computer vision~\cite{vpbmse-mcl-16,wdgh-auf-16,dpcn-lkc-16} and,
with the method described in this work, improvements to such predictive models could translate into improvements in robotic manipulation capabilities.
%%SL.9.15: add citations.

\section{Discussion \& Future Work}
\label{conclusion}

We presented a learning-based approach for basic nonprehensile manipulation. Our method uses a deep predictive neural network model to plan pushing motions with minimal prior engineering.
We use a convolutional LSTM model trained on unlabeled data collected autonomously by a team of robots. This model predicts future camera images and image-space pixel flow, conditioned on a sequence of motor commands.
By inferring the commands that will move individual points in an image to desired target locations, we can continuously plan for pushing tasks even with novel objects not previously seen during training.
Since the model is trained entirely through a self-supervised procedure, the method is well suited for continuous self-improvement through constant data collection.
Our experimental evaluation demonstrates that our method outperforms simple baselines based on geometric heuristics and known hand-to-camera calibration.

Although we show generalization to completely novel objects, our model is still limited to relatively simple short-horizon tasks.
As the accuracy of deep video prediction models improves, we expect the capabilities of this approach to also improve.
Predictive models that estimate distributions over future images are particularly promising for robotic control, since planning under such models corresponds to maximum likelihood inference for determining a sequence of actions that maximizes the probability of the desired outcome.
A promising direction for future work is to integrate the latest advances in such probabilistic video prediction models, e.g. ~\cite{wdgh-auf-16}, into our approach.
% I'm not sure there are other models that are probabilistic except pixel RNN (which isn't out yet) % maybe the video VAE paper from CMU

\addtolength{\textheight}{-0.4cm}   % This command serves to balance the column lengths

One of the enabling technologies for our approach is the available of fast, highly parallel GPUs for runtime evaluation of our model.
All of our experiments use a commercially available GPU, which makes the approach practical for self-contained robotic systems.
However, the planning horizon and replanning rate are limited by computational power, and the availability of even faster and more parallel computational platforms will likely lead to an improvement in capability and accuracy.

Finally, we believe that this work represents a relatively early first step toward model-based robotic control using learned predictive models.
Deep neural network-based video prediction is still in its early stages, with most state-of-the-art methods making accurate predictions only a few frames into the future~\cite{vpbmse-mcl-16}.  % TODO - is this SOTA?
As the state-of-the-art in video prediction improves, model-based methods will become increasingly more powerful.
Of particular interest for robotic manipulation, hierarchical models operating at varying time scales~\cite{hrnn-eb-95} may even make it practical to plan highly elaborate skills entirely using learned predictive models.

\section*{Acknowledgments} We thank Peter Pastor, Ethan Holly, Mrinal Kalakrishnan, Deirdre Quillen, and Stefan Hinterstoisser for technical support during the project, and Vincent Vanhoucke, Ian Goodfellow, and George Dahl for helpful discussions. We also thank Jon Barron for help with the optical flow solver.


%%%%%%%%%%%%%%%%%%%%%%%%%%%%%%%%%%%%%%%%%%%%%%%%%%%%%%%%%%%%%%%%%%%%%%%%%%%%%%%%

%%%%%%%%%%%%%%%%%%%%%%%%%%%%%%%%%%%%%%%%%%%%%%%%%%%%%%%%%%%%%%%%%%%%%%%%%%%%%%%%

%%%%%%%%%%%%%%%%%%%%%%%%%%%%%%%%%%%%%%%%%%%%%%%%%%%%%%%%%%%%%%%%%%%%%%%%%%%%%%%%
%\section*{APPENDIX}


%\section*{ACKNOWLEDGMENT}
%The preferred spelling of the word ÒacknowledgmentÓ in America is without an ÒeÓ after the ÒgÓ. Avoid the stilted expression, ÒOne of us (R. B. G.) thanks . . .Ó  Instead, try ÒR. B. G. thanksÓ. Put sponsor acknowledgments in the unnumbered footnote on the first page.

%%%%%%%%%%%%%%%%%%%%%%%%%%%%%%%%%%%%%%%%%%%%%%%%%%%%%%%%%%%%%%%%%%%%%%%%%%%%%%%%

\bibliographystyle{ieeetran}
\bibliography{references}

\begin{thebibliography}{10}
\providecommand{\url}[1]{#1}
\csname url@samestyle\endcsname
\providecommand{\newblock}{\relax}
\providecommand{\bibinfo}[2]{#2}
\providecommand{\BIBentrySTDinterwordspacing}{\spaceskip=0pt\relax}
\providecommand{\BIBentryALTinterwordstretchfactor}{4}
\providecommand{\BIBentryALTinterwordspacing}{\spaceskip=\fontdimen2\font plus
\BIBentryALTinterwordstretchfactor\fontdimen3\font minus
  \fontdimen4\font\relax}
\providecommand{\BIBforeignlanguage}[2]{{%
\expandafter\ifx\csname l@#1\endcsname\relax
\typeout{** WARNING: IEEEtran.bst: No hyphenation pattern has been}%
\typeout{** loaded for the language `#1'. Using the pattern for}%
\typeout{** the default language instead.}%
\else
\language=\csname l@#1\endcsname
\fi
#2}}
\providecommand{\BIBdecl}{\relax}
\BIBdecl

\bibitem{iv4-siv-16}
C.~Szegedy, S.~Ioffe, and V.~Vanhoucke, ``Inception-v4, inception-resnet and
  the impact of residual connections on learning,'' \emph{arXiv preprint
  arXiv:1602.07261}, 2016.

\bibitem{rfcn-dlhs-16}
J.~Dai, Y.~Li, K.~He, and J.~Sun, ``R-fcn: Object detection via region-based
  fully convolutional networks,'' \emph{arXiv preprint arXiv:1605.06409}, 2016.

\bibitem{ulpi-fgl-16}
C.~Finn, I.~Goodfellow, and S.~Levine, ``Unsupervised learning for physical
  interaction through video prediction,'' in \emph{Neural Information
  Processing Systems (NIPS)}, 2016.

\bibitem{osf-k-87}
O.~Khatib, ``A unified approach for motion and force control of robot
  manipulators: The operational space formulation,'' \emph{IEEE Journal on
  Robotics and Automation}, 1987.

\bibitem{mirm-mls-94}
R.~M. Murray, Z.~Li, and S.~S. Sastry, \emph{A mathematical introduction to
  robotic manipulation}.\hskip 1em plus 0.5em minus 0.4em\relax CRC press,
  1994.

\bibitem{pfnm-ds-12}
M.~R. Dogar and S.~S. Srinivasa, ``A planning framework for non-prehensile
  manipulation under clutter and uncertainty,'' \emph{Autonomous Robots}, 2012.

\bibitem{mpmp-m-86}
M.~Mason, ``Mechanics and planning of manipulator pushing operations,''
  \emph{The International Journal of Robotics Research (IJRR)}, 1986.

\bibitem{vbp-smos-93}
M.~Salganicoff, G.~Metta, A.~Oddera, and G.~Sandini, \emph{A vision-based
  learning method for pushing manipulation}.\hskip 1em plus 0.5em minus
  0.4em\relax AAAI Fall Symposium Series, 1993.

\bibitem{ppop-ches-11}
A.~Cosgun, T.~Hermans, V.~Emeli, and M.~Stilman, ``Push planning for object
  placement on cluttered table surfaces,'' in \emph{International Conference on
  Intelligent Robots and Systems (IROS)}, 2011.

\bibitem{mmwp-ybfr-16}
K.-T. Yu, M.~Bauza, N.~Fazeli, and A.~Rodriguez, ``More than a million ways to
  be pushed: A high-fidelity experimental data set of planar pushing,''
  \emph{International Conference on Intelligent Robots and Systems (IROS)},
  2016.

\bibitem{vlr3d-mn-95}
H.~Murase and S.~K. Nayar, ``Visual learning and recognition of 3-d objects
  from appearance,'' \emph{International Journal of Computer Vision (IJCV)},
  1995.

\bibitem{orfpr-cbsf-09}
A.~Collet, D.~Berenson, S.~S. Srinivasa, and D.~Ferguson, ``Object recognition
  and full pose registration from a single image for robotic manipulation,'' in
  \emph{International Conference on Robotics and Automation (ICRA)}, 2009.

\bibitem{ldd-etb-13}
F.~Endres, J.~Trinkle, and W.~Burgard, ``Learning the dynamics of doors for
  robotic manipulation,'' in \emph{International Conference on Intelligent
  Robots and Systems (IROS)}, 2013.

\bibitem{pp-mrd-03}
P.~McLeod, N.~Reed, and Z.~Dienes, ``Psychophysics: How fielders arrive in time
  to catch the ball,'' \emph{Nature}, 2003.

\bibitem{alvinn-p-89}
D.~Pomerleau, ``Alvinn; an autonomous land vehicle in a neural network,''
  \emph{Neural Information Processing Systems (NIPS)}, 1989.

\bibitem{llrv-hsbes-09}
R.~Hadsell, P.~Sermanet, J.~Ben, A.~Erkan, M.~Scoffier, K.~Kavukcuoglu,
  U.~Muller, and Y.~LeCun, ``Learning long-range vision for autonomous off-road
  driving,'' \emph{Journal of Field Robotics (JFR)}, 2009.

\bibitem{rghl-rlrs-09}
M.~Riedmiller, T.~Gabel, R.~Hafner, and S.~Lange, ``Reinforcement learning for
  robot soccer,'' \emph{Autonomous Robots}, 2009.

\bibitem{sss-pg-16}
L.~Pinto and A.~Gupta, ``Supersizing self-supervision: Learning to grasp from
  50k tries and 700 robot hours,'' \emph{International Conference on Robotics
  and Automation (ICRA)}, 2016.

\bibitem{lhecrg-lpkq-16}
S.~Levine, P.~Pastor, A.~Krizhevsky, and D.~Quillen, ``Learning hand-eye
  coordination for robotic grasping with deep learning and large-scale data
  collection,'' \emph{International Symposium on Experimental Robotics (ISER)},
  2016.

\bibitem{eetdvp-lfda-16}
S.~Levine, C.~Finn, T.~Darrell, and P.~Abbeel, ``End-to-end training of deep
  visuomotor policies,'' \emph{Journal of Machine Learning Research (JMLR)},
  2016.

\bibitem{dr-pilco-11}
M.~Deisenroth and C.~E. Rasmussen, ``Pilco: A model-based and data-efficient
  approach to policy search,'' in \emph{International Conference on Machine
  Learning (ICML)}, 2011.

\bibitem{rlhf-acqn-07}
P.~Abbeel, A.~Coates, M.~Quigley, and A.~Y. Ng, ``An application of
  reinforcement learning to aerobatic helicopter flight,'' \emph{Neural
  Information Processing Systems (NIPS)}, 2007.

\bibitem{tet-sscb-12}
Y.~Tassa, T.~Erez, and E.~Todorov, ``Synthesis and stabilization of complex
  behaviors through online trajectory optimization,'' in \emph{International
  Conference on Intelligent Robots and Systems (IROS)}, 2012.

\bibitem{dmpc-lks-15}
I.~Lenz, R.~Knepper, and A.~Saxena, ``Deepmpc: Learning deep latent features
  for model predictive control,'' in \emph{Robotics Science and Systems (RSS)},
  2015.

\bibitem{bbf-lpmdc-14}
B.~Boots, A.~Byravan, and D.~Fox, ``Learning predictive models of a depth
  camera \& manipulator from raw execution traces,'' in \emph{International
  Conference on Robotics and Automation (ICRA)}, 2014.

\bibitem{se3-bf-16}
A.~Byravan and D.~Fox, ``Se3-nets: Learning rigid body motion using deep neural
  networks,'' \emph{arXiv preprint arXiv:1606.02378}, 2016.

\bibitem{ltpbp-anaml-16}
P.~Agrawal, A.~Nair, P.~Abbeel, J.~Malik, and S.~Levine, ``Learning to poke by
  poking: Experiential learning of intuitive physics,'' \emph{Neural
  Information Processing Systems (NIPS)}, 2016.

\bibitem{bsg-psr-11}
B.~Boots, S.~M. Siddiqi, and G.~J. Gordon, ``Closing the learning-planning loop
  with predictive state representations,'' \emph{The International Journal of
  Robotics Research (IJRR)}, 2011.

\bibitem{lrv-arl-12}
S.~Lange, M.~Riedmiller, and A.~Voigtlander, ``Autonomous reinforcement
  learning on raw visual input data in a real world application,'' in
  \emph{International Joint Conference on Neural Networks (IJCNN)}, 2012.

\bibitem{e2c-wsbr-15}
M.~Watter, J.~Springenberg, J.~Boedecker, and M.~Riedmiller, ``Embed to
  control: A locally linear latent dynamics model for control from raw
  images,'' in \emph{Neural Information Processing Systems (NIPS)}, 2015.

\bibitem{ftddla-dsae-15}
C.~Finn, X.~Y. Tan, Y.~Duan, T.~Darrell, S.~Levine, and P.~Abbeel, ``Deep
  spatial autoencoders for visuomotor learning,'' \emph{International
  Conference on Robotics and Automation (ICRA)}, 2016.

\bibitem{ecr-navsr-92}
B.~Espiau, F.~Chaumette, and P.~Rives, ``A new approach to visual servoing in
  robotics,'' \emph{IEEE Transactions on Robotics and Automation}, 1992.

\bibitem{mkd-vbcqp-14}
K.~Mohta, V.~Kumar, and K.~Daniilidis, ``Vision based control of a quadrotor
  for perching on planes and lines,'' in \emph{International Conference on
  Robotics and Automation (ICRA)}, 2014.

\bibitem{whb-reecu-96}
W.~J. Wilson, C.~W.~W. Hulls, and G.~S. Bell, ``Relative end-effector control
  using cartesian position based visual servoing,'' \emph{IEEE Transactions on
  Robotics and Automation}, 1996.

\bibitem{bbm-cm-15}
A.~Censi and R.~M. Murray, ``{Bootstrapping bilinear models of Simple
  Vehicles},'' \emph{International Journal of Robotics Research}, vol.~34, July
  2015.

\bibitem{vpbmse-mcl-16}
M.~Mathieu, C.~Couprie, and Y.~LeCun, ``Deep multi-scale video prediction
  beyond mean square error,'' \emph{International Conference on Learning
  Representations (ICLR)}, 2016.

\bibitem{aevb-kw-13}
D.~P. Kingma and M.~Welling, ``Auto-encoding variational bayes,''
  \emph{International Conference on Learning Representations (ICLR)}, 2014.

\bibitem{ln-bkh-16}
J.~L. Ba, J.~R. Kiros, and G.~E. Hinton, ``Layer normalization,'' \emph{arXiv
  preprint arXiv:1607.06450}, 2016.

\bibitem{cem-rk-13}
R.~Y. Rubinstein and D.~P. Kroese, \emph{The cross-entropy method: a unified
  approach to combinatorial optimization, Monte-Carlo simulation and machine
  learning}.\hskip 1em plus 0.5em minus 0.4em\relax Springer Science \&
  Business Media, 2013.

\bibitem{jvrv-agbks-16}
R.~Anderson, D.~Gallup, J.~T. Barron, J.~Kontkanen, N.~Snavely, C.~Hern\'andez,
  S.~Agarwal, and S.~M. Seitz, ``Jump: Virtual reality video,'' \emph{SIGGRAPH
  Asia}, 2016.

\bibitem{wdgh-auf-16}
J.~Walker, C.~Doersch, A.~Gupta, and M.~Hebert, ``An uncertain future:
  Forecasting from static images using variational autoencoders,'' \emph{arXiv
  preprint arXiv:1606.07873}, 2016.

\bibitem{dpcn-lkc-16}
W.~Lotter, G.~Kreiman, and D.~Cox, ``Deep predictive coding networks for video
  prediction and unsupervised learning,'' \emph{arXiv preprint
  arXiv:1605.08104}, 2016.

\bibitem{hrnn-eb-95}
S.~El~Hihi and Y.~Bengio, ``Hierarchical recurrent neural networks for
  long-term dependencies.'' in \emph{Neural Information Processing Systems
  (NIPS)}, 1995.

\end{thebibliography}

\end{document}